\documentclass[sigconf, screen]{acmart}
\usepackage{wrapfig}
\usepackage{multirow}
\usepackage{makecell}
\usepackage{booktabs}
\usepackage{enumitem}
\usepackage{threeparttable}
\usepackage{pifont} % 导入 pifont 以使用 \ding
\usepackage[table]{xcolor}
\usepackage{xspace}  % 自动加空格
\newcommand{\ie}{\textit{i.e.}\xspace}
\newcommand{\eg}{\textit{e.g.}\xspace}

\AtBeginDocument{%
  }

\setcopyright{acmlicensed}
\copyrightyear{2026}
\acmYear{2026}
\acmDOI{XXXXXXX.XXXXXXX}
\acmConference[MM '26]{the 34th ACM International Conference on Multimedia}{November 10--14, 2026}{Janeiro, Brazil}

\acmISBN{978-1-4503-XXXX-X/2018/06}

\begin{document}

%\title{EgoVLA: Diffusion-Enhanced Vision-Language-Action Model for Ego-Centric Robotic Control}
%\title{EgoVLA: Towards Egocentric Robotic Control via Diffusion-Enhanced Vision-Language-Action Model}
% \title{LNCD: \underline{L}atent \underline{N}ovel-\underline{c}ategory Knowledge Self-\underline{D}istillation for Open-Vocabulary Object Detection} 
% \title{KDSD: Knowledge Discovery and Self-Distillation for Novel Category in Open-Vocabulary Object Detection}
\title{ExDet: Open-Domain Open-Vocabulary Detection with Cross-modal Extrapolation and Rectification}  
% 跨模态外推与重对齐的开放域开放词汇检测

\author{Yupeng Zhang}
\affiliation{%
  \institution{College of Intelligence and Computing, Tianjin University}
  \city{Tianjin}
  \country{China}}
\email{zhangyupeng@tju.edu.cn}
%%%%%%%%%%%%%%%%%%%%%%%%%%%
\author{Yuzhong Feng}
\affiliation{%
  \institution{College of Intelligence and Computing, Tianjin University}
  \city{Tianjin}
  \country{China}}
\email{yzfeng@tju.edu.cn}
%%%%%%%%%%%%%%%%%%%%%%%%%%%
\author{Ruize Han}
\affiliation{%
  \institution{Faculty of Computer Science and Artificial Intelligence, Shenzhen University of Advanced Technology}
  \city{Shenzhen}
  \country{China}}
\email{hanruize@suat-sz.edu.cn}
%%%%%%%%%%%%%%%%%%%%%%%%%%%
\author{Zhiwei Chen}
\affiliation{%
  \institution{School of Artificial Intelligence, Nanchang University}
  \city{Jiangxi}
  \country{China}}
\email{zhiweichen@ncu.edu.cn}
%%%%%%%%%%%%%%%%%%%%%%%%%%%
\author{Wei Feng}
\affiliation{%
  \institution{College of Intelligence and Computing, Tianjin University}
  \city{Tianjin}
  \country{China}}
\email{wfeng@ieee.org}
%%%%%%%%%%%%%%%%%%%%%%%%%%%
\author{Liang Wan}
\affiliation{%
  \institution{College of Intelligence and Computing, Tianjin University}
  \city{Tianjin}
  \country{China}}
\email{lwan@tju.edu.cn}

\begin{abstract}
% 开放域开放词汇检测（ODOVD）要求检测器同时泛化到新类别和未见领域，因此比传统开放词汇检测（OVD）更具挑战性。现有方法通常将开放词汇检测器与领域泛化模块从头联合训练，训练成本较高。为此，我们提出了 ExDet，一种面向 ODOVD 的轻量级类别--领域协同泛化框架，能够 detector training-free 和 real-data-free 的情况下增强现有检测器的跨类别与跨领域泛化能力。ExDet 由 Text-Guided Extrapolation（TGE）、轻量级 Detector-Compatible Rectification（DCR 检测器兼容校正）模块以及 ExRPN 组成。
% 具体而言，TGE 利用视觉语言模型（VLM）的 DeltaSpace 特性，从文本中推理出具备类别感知与领域感知的代理视觉原型。DCR 基于这些由 TGE 生成的原型，在独立于检测器训练的条件下进行学习，并在推理阶段接入分类头之后，将其表示重对齐到与检测器兼容的源域视觉分布，从而提升对新类别和未见领域目标的分类能力。ExRPN 则通过结合语义相似度与 RPN 置信度对 proposal 分数进行重校准，从而提高 novel 和 domain-shifted objects 的召回率，并为后续分类与 DCR 重对齐提供更好的候选支持。
% 在 OD-LVIS、OV-LVIS、Objects365 和 MSOSB 上的大量实验表明，ExDet 在类别与领域双重偏移条件下取得了最先进的性能。与此同时，ExDet 仅需在单张 NVIDIA RTX 3090 上训练约 30 分钟，具有较高的训练效率。
Open-domain open-vocabulary detection (ODOVD) requires detectors to generalize to both novel categories and unseen domains, making it more challenging than open-vocabulary detection. Existing methods typically train open-vocabulary detectors together with domain generalization modules from scratch, leading to high training cost. 
we propose ExDet, a lightweight category-domain collaborative generalization framework for ODOVD that enhances the cross-category and cross-domain generalization of existing detectors.
ExDet consists of Text-Guided Extrapolation (TGE), a lightweight Detector-Compatible Rectification (DCR) module, and ExRPN.
Specifically, TGE exploits the DeltaSpace property of vision-language models (VLMs) to infer category- and domain-aware proxy visual prototypes from text. 
DCR is learned from the TGE-generated prototypes  in a detector training-free and real-data-free manner, and is inserted after the classification head at inference to rectify representations toward a detector-compatible source-domain visual distribution, thereby enhancing classification for targets from novel categories and unseen domains. 
ExRPN recalibrates proposal scores by combining semantic similarity with RPN confidence, improving recall for novel and domain-shifted objects while providing better support for subsequent classification and DCR.
ExDet achieves SOTA performance on OD-LVIS, OV-LVIS, Objects365, and MSOSB. 
% while requiring only about 30 minutes of training on a single RTX 3090 GPU.
\end{abstract}




\maketitle

\section{Introduction}
\label{sec:intro}

\begin{figure}[t!]
	\centering
	\includegraphics[width=\linewidth]{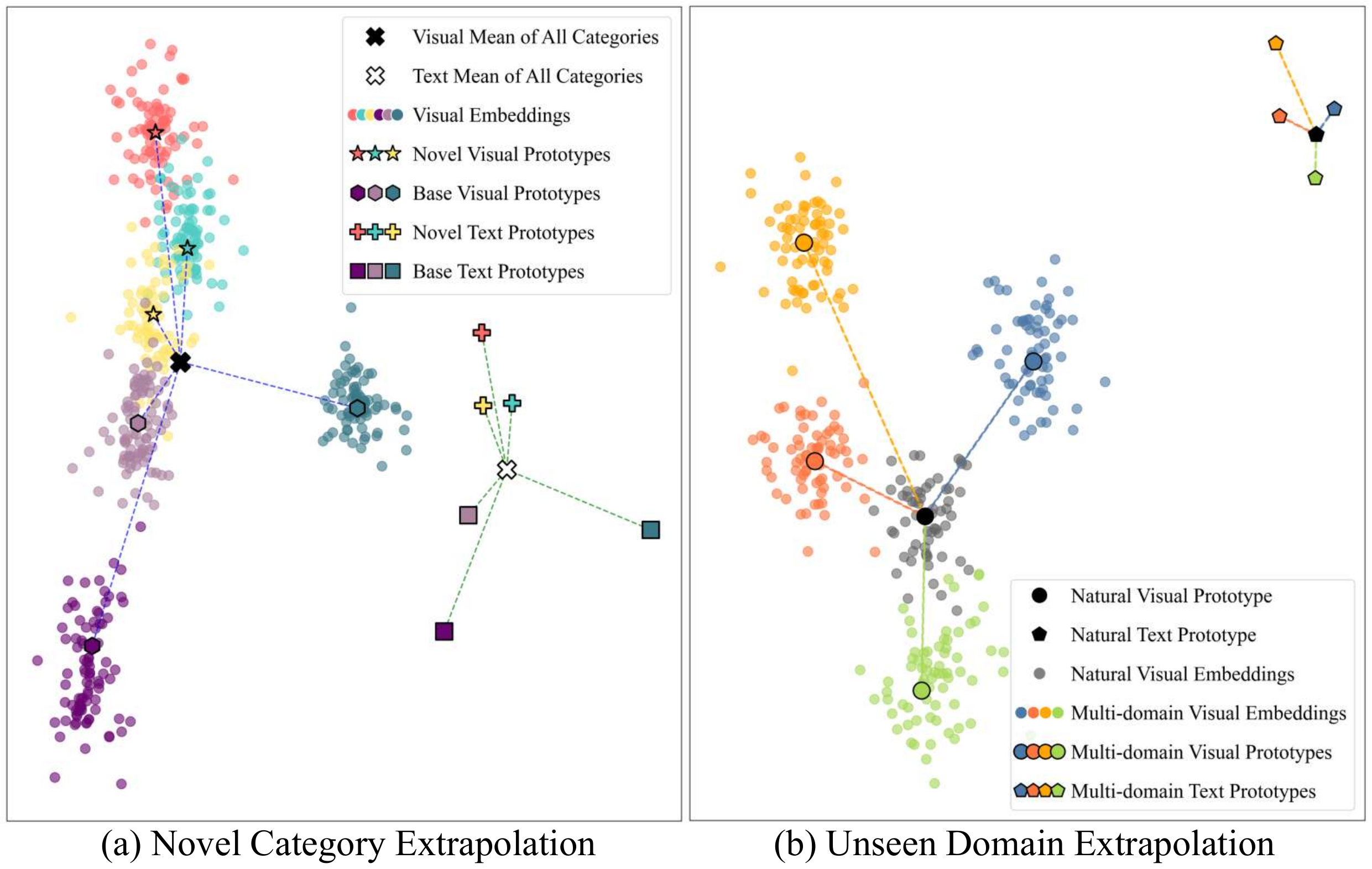}	
        \vspace{-20pt}
        \caption{PCA visualization of visual extrapolation for novel categories and unseen domains, where text and image features are slightly shifted in the shared representation space for clarity. (a) Novel Category Extrapolation: The relative distribution of base and novel categories in the text space is consistent with that of their corresponding visual prototypes in the visual space. (b) Unseen Domain Extrapolation: For the same category, the offset from multi-domain text prototypes to the natural text prototype corresponds structurally to the offset from the corresponding visual prototypes to the natural-domain visual prototype.}  % 用于新类别和未见域视觉外推的 PCA 可视化。 为了便于观察，在共享表征空间中对文本特征和图像特征进行了轻微平移。(a) 新类别外推： 文本空间中基础类别与新类别的相对分布关系，与视觉空间中对应视觉原型的相对分布关系保持一致。(b) 未见域外推： 对于同一类别，从多域文本原型到自然域文本原型的偏移，在结构上对应于从相应视觉原型到自然域视觉原型的偏移。
	\label{fig:distribute}
	\vspace{-15pt}
\end{figure}

% 近年来，目标检测在监督学习框架下取得了显著进展，但现有方法大多依赖大规模人工标注，并建立在预定义闭集类别之上，因而难以满足真实场景中持续扩展的开放类别需求。为此，开放词汇检测（Open-Vocabulary Detection, OVD）~\cite{zareian2021open}借助视觉语言模型（Vision-Language Models, VLMs）~\cite{radford2021learning,jia2021scaling}的开放语义表征能力，使检测器能够从有限标注类别泛化到未见类别。然而，现实应用中的分布变化通常不仅表现为类别偏移，还常伴随着视觉域偏移~\cite{lin2021domain}，从而形成更具挑战性的联合类别-域偏移。比如，当检测器从室内场景迁移到室外环境时，目标类别、光照、天气及整体成像风格往往会同时变化。为应对这一挑战，Open-Domain Open-Vocabulary Detection (ODOVD)~\cite{zhang2025odov} 被提出，旨在在开放词汇设定下同时实现跨类别与跨领域的统一目标检测泛化。
In recent years, object detection has achieved remarkable progress under supervised learning. However, existing methods largely rely on large-scale manual annotations and assume a predefined closed-set label space, making them inadequate for real-world scenarios with ever-expanding object categories. To address this limitation, Open-Vocabulary Detection (OVD)~\cite{zareian2021open} leverages the open semantic representations of vision-language models (VLMs)~\cite{radford2021learning,jia2021scaling} to generalize from limited annotated categories to unseen ones.
However, distribution shifts in real applications often go beyond category shift alone and are frequently accompanied by visual domain shift~\cite{lin2021domain}, resulting in the more challenging setting of joint category-domain shift. For example, when a detector trained on indoor scenes is deployed in outdoor environments, object categories, illumination, weather, and overall imaging style may all change simultaneously. To tackle this challenge, Open-Domain Open-Vocabulary Detection (ODOVD)~\cite{zhang2025odov} has been introduced to achieve unified object detection generalization across both categories and domains.

% 尽管已有 ODOVD 方法 DVtor~\cite{zhang2025odov} 通过引入域嫁接机制来提升模型的跨域鲁棒性，但其仍需与检测框架共同训练，带来较高的训练成本与系统复杂度。另外，DVtor 主要聚焦于分类判别能力的增强，而对两阶段检测器中的候选区域生成问题关注不足。事实上，Region Proposal Network (RPN) 仅基于源域基类标注进行训练，当面对新类别或域偏移目标时，往往难以生成高质量、高置信度的 proposals，导致大量潜在目标在后处理阶段即被提前过滤。由此，ODOVD 的挑战不仅在于提升分类头对 novel categories 和 unseen domains 的识别能力，也在于提高 proposal 阶段对 novel and domain-shifted objects 的召回率，从而为后续 RoI head 提供更充分的候选区域支持。
Although DVtor~\cite{zhang2025odov}, the pioneering work on ODOVD, improves cross-domain robustness by introducing a domain grafting mechanism, it requires joint training with the detector, leading to increased training cost and system complexity. 
In fact, DVtor mainly focuses on enhancing classification capability, while paying limited attention to proposal generation in two-stage detectors. In practice, the Region Proposal Network (RPN) is trained only on base-category annotations from the source domain, and thus often struggles to produce high-quality, high-confidence proposals for novel or domain-shifted objects. As a result, many latent objects are filtered out before reaching the RoI head. Therefore, the challenge of ODOVD lies not only in improving classification for novel categories and unseen domains, but also in enhancing proposal recall for novel and domain-shifted objects.

% 我们观察到，VLMs 中的文本空间与视觉空间并非彼此孤立，而是呈现出一定的结构对应关系。特别地，DeltaSpace~\cite{lyu2023deltaedit,lyu2023deltaspace} 表明，文本语义空间中的方向性变化在一定程度上能够映射为视觉空间中的语义偏移。尽管已有研究~\cite{chen2025trtst,liu2025s,wang2023text,wu2023mfeclip} 在图像编辑及相关任务中验证了 DeltaSpace 的有效性，但其在目标检测中的适用性仍有待充分探索。进一步地，如图~\ref{fig:distribute} 所示，我们的初步可视化分析表明，在目标概念层面，类别与领域变化所引起的文本语义偏移与视觉空间中的特征偏移均呈现出较为一致的结构关系。
% 基于这一观察，我们自然会问：能否仅依赖类别与领域描述，推理得到新类别与未见领域的代理视觉原型，并进一步设计一个轻量级兼容模块，以 detector training-free and real-data-free 的方式提升检测器的跨类别与跨领域泛化能力？
We observe that the text and visual spaces of VLMs are not isolated, but exhibit structural correspondence. In particular, DeltaSpace~\cite{lyu2023deltaedit,lyu2023deltaspace} suggests that directional changes in the text semantic space can translate into semantic shifts in the visual space. Although prior studies~\cite{chen2025trtst,liu2025s,wang2023text,wu2023mfeclip} have validated the effectiveness of DeltaSpace in image editing and related tasks, its applicability to object detection remains underexplored. Furthermore, as shown in Fig.~\ref{fig:distribute}, our preliminary visualization reveals that, \textbf{\textit{at the object-concept level, category- and domain-induced text semantic shifts exhibit a broadly consistent structural relationship with feature shifts in the visual space}}.
Motivated by this observation, we ask a key question: \textit{can we infer proxy visual prototypes for novel categories and unseen domains solely from category and domain descriptions, and design a lightweight compatibility module to enhance cross-category and cross-domain generalization in a detector training-free and real-data-free manner?}

% 基于上述观察，本文提出一种用于开放域开放词汇检测的轻量级增强框架 ExDet，通过文本引导的视觉外推、分布兼容重对齐和推理期 proposal 校正，同时缓解两阶段检测器在联合类别-域偏移下的两个关键瓶颈：1) 分类头对新类别与新领域目标的判别不足；2) proposal 阶段对 novel/domain-shifted objects 的召回不足。在无需重新训练检测器的前提下，该框架能够有效提升其跨类别与跨领域泛化能力。
% 具体而言，针对前者，我们提出 Text-Guided Extrapolation (TGE) 和 Detector-Compatible Rectification (DCR)。其中，TGE 利用文本嵌入在 VLM 特征空间中的分布关系，外推出跨类别、跨领域的代理视觉原型；DCR 则以这些增强原型为监督进行独立训练，而且detector training-free 和 real-data-free，并在推理阶段将分类头后的表示重对齐到与检测器兼容的源域视觉分布，从而提升对新类别和新领域目标的分类判别能力。针对后者，我们进一步引入一种补充性的推理期策略 ExRPN，对 proposal 置信度进行语义感知校正，以提高 novel/domain-shifted objects 的召回率，并为后续分类与校准提供更充分的候选支持。
Based on the above observations, we propose ExDet, a lightweight enhancement framework for ODOVD. By combining text-guided visual extrapolation, Detector-Compatible Rectification, and inference-time proposal rectification, ExDet addresses two key bottlenecks of two-stage detectors under joint category--domain shifts: 
1) insufficient discrimination of the classification head for novel categories and unseen domains, and 
2) limited proposal recall for novel and domain-shifted objects. Without retraining the detector, ExDet effectively improves both cross-category and cross-domain generalization.
Specifically, to address the former, we introduce \textbf{Text-Guided Extrapolation (TGE)} and \textbf{Detector-Compatible Rectification (DCR)}. 
TGE exploits the distributional relationships of text embeddings in the VLM feature space to extrapolate cross-category, cross-domain proxy visual prototypes. 
DCR is independently trained under the supervision of enhanced prototypes and, in a \textbf{detector training-free} and \textbf{real-data-free} manner, rectifies post-classification-head representations toward a detector-compatible source-domain visual distribution, thereby improving discrimination for objects from novel categories and unseen domains.
To address the latter, we further introduce ExRPN, a complementary inference-time strategy that semantically rectifies proposal confidence scores to improve recall for novel and domain-shifted objects while providing better candidate support for subsequent classification and rectification.

In summary, the main contributions of this paper are as follows:
\begin{itemize}[topsep=0pt, itemsep=0pt, parsep=0pt]
\item We propose ExDet, a lightweight framework for ODOVD that enhances the cross-category and cross-domain generalization of existing two-stage detectors under joint category–domain shifts in a detector training-free and real-data-free manner.

\item We design a framework consisting of Text-Guided Extrapolation (TGE), a lightweight Detector-Compatible Rectification (DCR) module trained independently of the detector, and ExRPN. TGE constructs category- and domain-aware proxy visual prototypes from text descriptions; DCR rectifies classification representations toward a detector-compatible source-domain visual distribution for better discrimination of novel categories and unseen domains; and ExRPN recalibrates proposal confidence at inference to improve recall of novel and domain-shifted objects while supporting subsequent classification and rectification.

\item Extensive experiments on OD-LVIS~\cite{zhang2025odov}, OV-LVIS~\cite{gupta2019lvis}, Objects365~\cite{shao2019objects365}, and MSOSB~\cite{zhang2024rethinking} demonstrate that our method achieves SOTA performance and strong generalization under both category and domain shifts. ExDet also trains in only about 30 minutes on a single RTX 3090 GPU.
\end{itemize}

\section{Related Work}
\label{sec:Related work}
\textbf{Open-Vocabulary Detection (OVD).} With the rapid development of VLMs such as CLIP~\cite{radford2021learning} and ALIGN~\cite{jia2021scaling}, OVD~\cite{zareian2021open} has become an important research direction, aiming to recognize both base and novel categories within a unified cross-modal semantic space. Existing methods mainly rely on knowledge distillation, large-scale region-text supervision, transfer learning, or pseudo-labeling. For example, ViLD~\cite{gu2021open}, DetPro~\cite{du2022learning}, OADP~\cite{wang2023object}, and DK-DETR~\cite{li2023distilling} distill CLIP knowledge into detectors, while RO-ViT~\cite{kim2023region}, CORA~\cite{wu2023cora}, YOLO-World~\cite{cheng2024yolo}, and YOLOE~\cite{wang2025yoloe} improve open-world perception with large-scale region-text data, albeit at high cost. Pseudo-labeling methods, such as Detic~\cite{zhou2022detecting}, OCO~\cite{bangalath2022bridging}, ProxyDet~\cite{jeong2024proxydet}, LBP~\cite{li2024learning}, SAS-Det~\cite{zhao2024taming}, and OV-DQUO~\cite{wang2025ov}, mine potential objects from image-level tags, class-agnostic detectors, or proxy categories, but their reliance on text matching often introduces noise and limits generalization to novel categories. Meanwhile, F-VLM~\cite{kuo2022f}, CLIPSelf~\cite{wu2023clipself}, DST-Det~\cite{xu2024dst}, NoOVD~\cite{zhang2026noovdnovelcategorydiscovery}, and DeCLIP~\cite{wang2025declip} build two-stage detectors on frozen CLIP models, training only the detection heads for open-vocabulary recognition.
As the first ODOVD method, DVtor~\cite{zhang2025odov} also builds on frozen VLMs, better preserving the category- and domain-generalization ability inherited from VLM pre-training. 
Building on this paradigm, we introduce a Detector-Compatible Rectification (DCR) module after the classification head to improve inference-time generalization to novel categories and unseen domains, together with ExRPN to recalibrate proposal confidence scores and enhance recall for novel and domain-shifted objects.

\begin{figure*}[t!]
	\centering
	\includegraphics[width=1.0\linewidth]{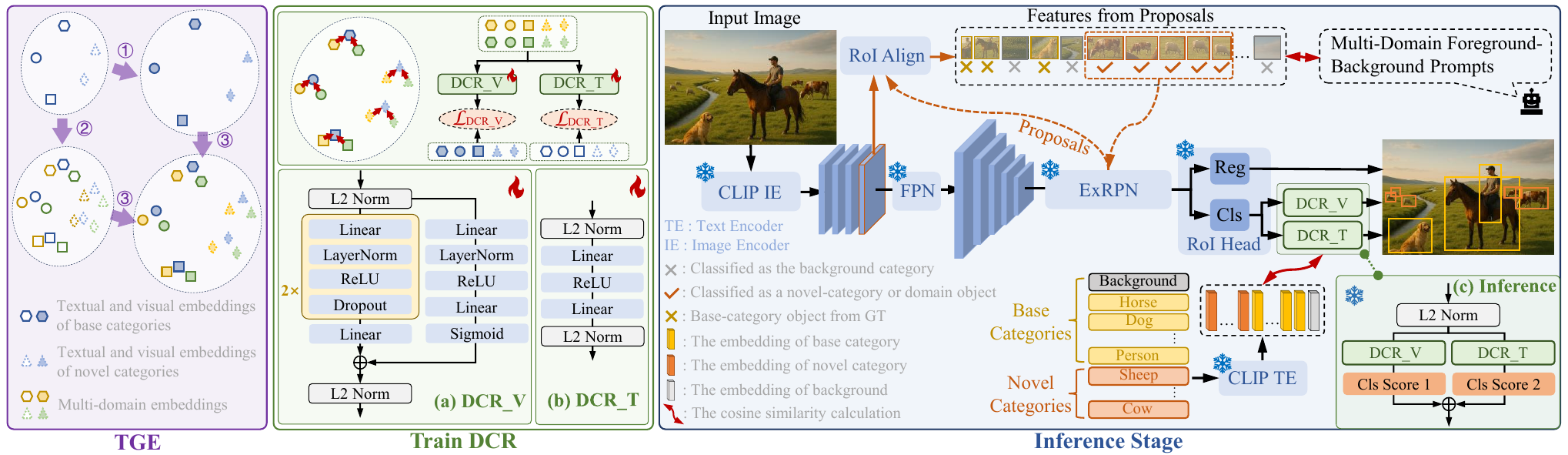}
	\vspace{-22pt}
	\caption{Overview of ExDet. Built upon a frozen F-ViT detector, ExDet is a lightweight category--domain generalization framework consisting of Text-Guided Extrapolation (TGE), Detector-Compatible Rectification (DCR), and ExRPN. TGE (\ding{172}--\ding{174}) generates enhanced embeddings covering novel categories and diverse domains to train DCR independently, where DCR comprises a visual branch (DCR\_V) for source-domain visual alignment and a semantic branch (DCR\_T) for preserving compatibility with the text space.
    During inference, DCR is inserted after the classification head to improve category--domain generalization. To further enhance recall for novel and domain-shifted objects, ExRPN recalibrates proposal confidence scores using coarse-grained semantic prompts adopted from NoOVD.}
    % ExDet 方法概览。ExDet 构建于冻结的 F-ViT 检测器之上，是一个轻量级的类别--领域泛化框架，由 Text-Guided Extrapolation（TGE）、Detector-Compatible Rectification（DCR）和 ExRPN 三部分组成。TGE（\ding{172}--\ding{174}）生成覆盖新类别和多样领域的增强嵌入，并用于独立训练 DCR，其中 DCR 包含一个用于源域视觉对齐的视觉分支（DCR_V）和一个用于保持文本空间兼容性的语义分支（DCR_T）。在推理阶段，DCR 被插入到分类头之后，以提升类别--领域泛化能力。为进一步增强对新类别和域偏移目标的召回率，ExRPN 利用从 NoOVD 引入的粗粒度语义提示对 proposal 置信度进行重校准。
	\label{fig:overview}
	\vspace{-10pt}
\end{figure*}

\textbf{Domain Generalization (DG) based object detection} aims to train detectors on source domains that generalize to unseen target domains. Early works explored feature disentanglement for cross-domain generalization~\cite{lin2021domain}, and Gated Disentangling Network~\cite{zhang2022gated} further improved generalization by activating domain-invariant channels. However, these methods rely on data and annotations from multiple source domains. In contrast, single-domain generalized object detection (SDGOD)~\cite{wu2022single} considers the more challenging setting where only one source domain is available.
Existing SDGOD methods can be broadly categorized into four groups. Data augmentation methods, such as CLIP the Gap~\cite{vidit2023clip}, DivAlign~\cite{danish2024improving}, and SRCD~\cite{rao2024srcd}, improve robustness by expanding the training distribution with image- or feature-level perturbations, but often still rely on two-stage detectors or auxiliary modules. Feature disentanglement methods, including SDGOD~\cite{wu2022single}, DG-DETR~\cite{hwang2025dg}, and UFR~\cite{liu2024unbiased}, separate domain-invariant and domain-specific factors through dedicated architectures or objectives, yet usually introduce additional computation or inference branches. Architecture search methods, such as G-NAS~\cite{wu2024g}, seek detector structures more favorable for generalization, while test-time adaptation methods, \eg, SA-DETR~\cite{han2025style}, adapt models to unseen domains through dynamic inference-time adjustment, at the cost of extra overhead.
Our setting is closely related to SDGOD, with training on a single source domain and evaluation across multiple open domains. Unlike conventional DG-based detectors, we address the more challenging ODOVD setting, which demands simultaneous generalization to novel categories and unseen domains.

\textbf{Semantic DeltaSpace of VLMs.} VLMs, especially CLIP~\cite{radford2021learning}, have shown strong cross-modal representation and alignment ability by learning a unified semantic space from large-scale image-text pairs. Recent studies~\cite{lyu2023deltaspace,chen2025trtst,liu2025s,wang2023text,wu2023mfeclip,lyu2023deltaedit} show that, in the semantic DeltaSpace of VLMs, shifts in text embeddings can partially induce corresponding shifts in visual embeddings. 
Inspired by this, we propose TGE to extrapolate proxy visual features for novel categories and unseen domains from textual semantic relations, together with a lightweight DCR module that rectifies them toward a detector-compatible source-domain visual distribution, thereby improving generalization to targets from novel categories and unseen domains in a detector training-free and real-data-free manner.

%%%%%%%%%%%%%%%%%%%%%%%%%%%%%%%%%%%%%%%%%%%%%%%%%%%%%%%%%%%%%%%%%%%%%%%%%%%%%%%%%%%%%%%%%%%%%%%%%%%%%%%%%%%%%%%%%%%%%%%%%%%%%%%%%%%%%%%%%%%%%%%%%%%%%%%%%%%%%%%%%%%%%%%%%
\section{Proposed Method}
\label{sec:Method}
\subsection{Preliminaries and Framework}

\textbf{Problem formulation.}
During the training stage, we use the data from the \textit{single source domain} (\ie, the natural scenes). Specifically, the training image dataset is notated as $\mathcal{D}^\mathrm{train} = \{\mathbf{I}_i^\mathrm{train}, L_i\}_{i=1}^{N}$, where $N$ is the number of training images, $\mathbf{I}_i$ denotes an image from the source domain with the detection annotation $L_i$ for it.
The label $L_i$ is composed of $\{\mathbf{b}_i, \mathbf{c}_i\}$, in which $\mathbf{b}_i$ indicates all the annotated object bounding boxes in $\mathbf{I}_i$, and the corresponding object categories are stored in $\mathbf{c}_i$.
Note that, all the (annotated) object categories contained in $L_i$ of the training set $\mathcal{D}^\mathrm{train}$ are from the \textit{base category} set, \ie, $\mathcal{C}^\mathrm{base}$.
During the testing stage, the ODOVD task requires the model to operate under both \textit{open-domain} and \textit{open-vocabulary} conditions. Specifically, the test images are sampled from hybrid open domains (\ie, exhibiting diverse image styles), denoted as $\mathcal{D}^\mathrm{test} = \{\mathbf{I}_j^\mathrm{test}\}_{j=1}^{M}$, where $M$ is the dataset scale.
For each test image $\mathbf{I}_j^\mathrm{test}$, the model is expected to output a set of predicted object bounding boxes and their corresponding categories, \ie, $\{\mathbf{b}_j, \mathbf{c}_j\}$.
Following the OVD setting, the predicted categories include both the \textit{base categories} $\mathcal{C}^\mathrm{base}$ (seen during training) and the \textit{novel categories} $\mathcal{C}^\mathrm{novel}$ (unseen during training), which together form the \textit{open-vocabulary category set}, defined as: $\mathcal{C}^\mathrm{open} = \mathcal{C}^\mathrm{base} + \mathcal{C}^\mathrm{novel}$.

\textbf{Overview.}
In Fig.~\ref{fig:overview}, we propose ExDet, a lightweight category--domain co-generalization framework built upon a frozen two-stage open-vocabulary detector. 
It consists of three key components: Text-Guided Extrapolation (TGE), Detector-Compatible Rectification (DCR), and ExRPN. 
Specifically, TGE exploits the DeltaSpace property of VLMs to extrapolate category-aware and domain-aware proxy visual prototypes, while DCR is independently trained under the supervision of these enhanced prototypes and plugged into the detector at inference to rectify post-classification-head region representations toward a detector-compatible source-domain visual distribution, thereby improving classification of targets from novel categories and unseen domains.
As a complementary inference-time strategy, ExRPN recalibrates proposal scores based on the similarity between proposal features and multi-domain semantic descriptions, improving recall for novel and domain-shifted objects while providing better candidate support for subsequent classification and DCR. 
Overall, ExDet improves the cross-category and cross-domain generalization ability of existing detectors in a \textbf{detector training-free} and \textbf{real-data-free} manner.
% 如图~\ref{fig:overview} 所示，我们提出了一种基于冻结两阶段开放词汇检测器的轻量级类别--领域协同泛化框架ExDet，由 Text-Guided Extrapolation (TGE)、Detector-Compatible Rectification (DCR) 和 ExRPN 三个关键组件组成。
% 其中，TGE 利用 VLM 的 DeltaSpace 特性外推出具备类别感知与领域感知的代理视觉原型；DCR 则以这些增强原型为监督独立训练，并于推理阶段接入检测器，对候选区域经过分类头后的表示进行重校正，使其回归到与检测器兼容的源域视觉分布，从而提升对新类别与未见领域目标的分类能力。
% 作为一种补充性的推理期策略，ExRPN 基于 proposal 特征与多领域语义描述之间的相似性对 proposal 分数进行重校准，从而提升新类别与域偏移目标的召回率，并为后续分类与 DCR 提供更充分的候选区域支持。
% 总体而言，该框架在无需重新训练现有开放词汇检测器的前提下，有效增强了其跨类别与跨领域泛化能力。

\subsection{Text-Guided Extrapolation}
\label{sec:TGE}
Since the training data cover only base categories in a single source domain, the detector lacks direct exposure to the visual variations of novel categories and unseen domains. To address this limitation without introducing additional real data, we propose \textbf{Text-Guided Extrapolation (TGE)} in the semantic space of VLMs, as shown on the left of Fig.~\ref{fig:overview}. By exploiting the DeltaSpace~\cite{lyu2023deltaedit,lyu2023deltaspace} property of VLMs, TGE synthesizes proxy visual prototypes for both novel categories and unseen domains.

\textbf{Novel category prototype extrapolation.}
We first encode all test categories using a fixed text template with a pre-trained VLM (\eg, CLIP). For base categories, we extract the corresponding visual embeddings from the frozen detector's classification head, normalize them, and average them to obtain a base-category visual prototype. We then transfer the offset between the novel-category text prototype and the base-category text prototype into the visual space to obtain the visual prototype of each novel category.

Specifically, let $t_n$ denote the text prototype of novel category $n$, and $\bar{t}_b$ and $\bar{v}_b$ the base-category text and visual prototypes. The extrapolated visual prototype for category $n$ is defined as
\begin{equation}
\label{eq:1}
\footnotesize
\hat{v}_n = \bar{v}_b + (t_n - \bar{t}_b).
\end{equation}
This linear extrapolation requires no additional training and provides zero-shot proxy visual prototypes for novel categories.

\textbf{Domain-aware text prototype construction.}
To simulate visual domain variations commonly encountered in real-world scenarios, we construct a diverse set of domain labels $\{D_j\}_{j=1}^N$ (\eg, \textit{`rainy'}, \textit{`blurry'}, \textit{`oil painting'}). These labels are combined with category names under a unified template, \textit{`a [domain] image of a/an [category]'}, to form multi-domain textual descriptions. We then encode these descriptions with a pre-trained VLM (\eg, CLIP) to obtain the corresponding domain-aware text prototypes, denoted as $t^D_{c,j}$, where $c$ indexes the category (including both base and novel categories) and $j$ indexes the domain type.

\textbf{Cross-domain visual prototype generation.}
Moreover, we synthesize domain-aware visual prototypes by combining each category's visual prototype with the relative shift between its domain-specific and general text prototypes. Similar to category extrapolation, we assume that the offset between a category's domain-specific text prototype and its general text prototype, \ie, $\Delta_{c,j}^D = t^D_{c,j} - t_c$, can be approximately transferred to the visual space. By adding this offset to the visual prototype of category $c$, we obtain its domain-aware visual prototype under domain $j$:
\begin{equation}
\footnotesize
\hat{\boldsymbol{v}}_{c,j}^{D} = \hat{\boldsymbol{v}}_c + \Delta_{c,j}^D,
\end{equation}
where $\hat{\boldsymbol{v}}_{c,j}^{D}$ denotes the domain-aware visual prototype of category $c$ under domain $j$, $t_c$ denotes the general text prototype of category $c$, and $\hat{\boldsymbol{v}}_c$ denotes the category visual prototype obtained from Eq.~(\ref{eq:1}), including those synthesized for novel categories. \textit{To further enrich diversity, we additionally construct mixed-domain visual prototypes through randomized weighted combinations of different domain-aware visual prototypes}.
% \textbf{跨域视觉原型生成.}
% 此外，我们通过将每个类别的视觉原型与其领域特定文本原型和通用文本原型之间的相对偏移相结合，来合成领域感知的视觉原型。类似于类别外推，我们假设类别的领域特定文本原型与通用文本原型之间的偏移，即 $\Delta_{c,j}^D = t^D_{c,j} - t_c$，可以近似迁移到视觉空间中。通过将该偏移加到类别 $c$ 的视觉原型上，我们可以得到其在领域 $j$ 下的领域感知视觉原型：
% \begin{equation}
% \footnotesize
% \hat{\boldsymbol{v}}_{c,j}^{D} = \hat{\boldsymbol{v}}_c + \Delta_{c,j}^D,
% \end{equation}
% 其中，$\hat{\boldsymbol{v}}_{c,j}^{D}$ 表示类别 $c$ 在领域 $j$ 下的领域感知视觉原型，$t_c$ 表示类别 $c$ 的通用文本原型，$\hat{\boldsymbol{v}}_c$ 表示由式~(\ref{eq:1}) 得到的类别视觉原型，其中也包括为新类别合成的视觉原型。 \textit{为进一步丰富多样性，我们还通过对不同领域感知视觉原型进行随机加权组合，额外构造了混合域视觉原型。}

%%%%%%%%%%%%%%%%%%%%%%%%%%%%%%%%%%%%%%%%%%%%%%%%%%%%%%%%%%%%%%%%%%%%%%%%%%%%%%%%%%%%%%%%%%%%%%%%%%%%%%%%%%%%%%%%%%%
\subsection{Detector-Compatible Rectification}  % 原来 Distribution-Compatible Re-alignment (DCR)  
\label{sec:DCR}
% TGE 为检测器构造了具备类别感知与领域感知的增强视觉原型，从而在不引入额外真实数据的前提下补足了新类别与未见域的视觉变化。然而，这些由文本引导外推得到的合成特征并不天然与原始检测器的分类空间兼容。为使检测器能够有效利用这些增强原型，我们进一步利用其作为监督信号来训练一个轻量级的\textbf{检测器兼容校正（Detector-Compatible Rectification, DCR）}网络；在推理阶段，DCR 作用于候选区域经过分类头后的表示，并在保持类别判别性的同时，将其重校正到与检测器兼容的源域视觉分布。如图~\ref{fig:overview} 中间部分所示，DCR 由两个互补分支组成：视觉分支（\textbf{DCR_V}）用于源域原型对齐，文本分支（\textbf{DCR_T}）用于保持与冻结文本分类器的语义一致性。
TGE constructs category- and domain-aware augmented visual prototypes, thereby compensating for the missing visual variations of novel categories and unseen domains \textbf{without introducing additional real data}. However, these text-guided synthesized features are not naturally compatible with the original detector's classification space. To make them effectively usable, we employ them as supervision to train a lightweight \textbf{Detector-Compatible Rectification (DCR)} network; during inference, DCR operates on the post-classification-head representations of candidate regions and rectifies them to the detector-compatible source-domain visual distribution while preserving category discriminability. As shown in the middle of Fig.~\ref{fig:overview}, DCR consists of two complementary branches: a visual branch (\textbf{DCR\_V}) for source-domain prototype rectification and a text branch (\textbf{DCR\_T}) for semantic consistency with the frozen text classifier.

\textbf{Visual branch: DCR\_V.}
As illustrated in Fig.~\ref{fig:overview} (a), DCR\_V adopts a gated residual architecture to rectify the TGE-generated domain-aware visual prototypes to the corresponding source-domain category prototypes. Given an input prototype $\hat{\boldsymbol{v}}_{c,j}^{D}$, DCR\_V outputs a refined representation $\tilde{\boldsymbol{v}}_{c,j}^{D,v}$.

The main branch of DCR\_V consists of two stacked feed-forward layers that model nonlinear distribution shifts, while an auxiliary gating branch produces a modulation vector to adaptively control the transformation. The outputs of the two branches are combined through residual addition, allowing DCR\_V to preserve category identity while correcting domain-induced deviations.

The DCR\_V is trained with three complementary objectives:

\textit{Cosine-based cross-entropy loss.}
This loss encourages the refined prototype to be discriminatively aligned with the correct source-domain category prototype:
\begin{equation}
\label{eq:ce}
\footnotesize
\mathcal{L}_{\mathrm{CE}}^{(v)}
=
-\log
\frac{
\exp\!\left(\cos(\tilde{\boldsymbol{v}}_{c,j}^{D,v}, \hat{\boldsymbol{v}}_{c})/\tau\right)
}{
\sum\limits_{c'}
\exp\!\left(\cos(\tilde{\boldsymbol{v}}_{c,j}^{D,v}, \hat{\boldsymbol{v}}_{c'})/\tau\right)
},
\end{equation}
where $\hat{\boldsymbol{v}}_{c'}$ denotes the source-domain visual prototypes of all categories, and $\tau$ is a temperature hyperparameter.

\textit{L2 reconstruction loss.}
To minimize the distance between the refined prototype and its source-domain counterpart, we use
\begin{equation}
\label{eq:l2}
\footnotesize
\mathcal{L}_{\mathrm{L2}}^{(v)}
=
\left\|
\tilde{\boldsymbol{v}}_{c,j}^{D,v} - \hat{\boldsymbol{v}}_{c}
\right\|_2^2.
\end{equation}

\textit{Contrastive loss.}
To improve inter-category separability and reduce feature confusion, we adopt a supervised contrastive objective:
\begin{equation}
\label{eq:cons}
\footnotesize
\mathcal{L}_{\mathrm{con}}^{(v)}
=
\sum_{i \in \mathcal{I}}
\frac{-1}{|\mathcal{P}(i)|}
\sum_{p \in \mathcal{P}(i)}
\log
\frac{
\exp\!\left(z_i \cdot z_p/\tau\right)
}{
\sum_{a \in \mathcal{A}(i)}
\exp\!\left(z_i \cdot z_a/\tau\right)
},
\end{equation}
where $\mathcal{I}$ denotes the index set of all samples in a batch, $\mathcal{P}(i)$ is the set of positive samples sharing the same label as anchor $i$, $\mathcal{A}(i)$ is the set of all comparison samples for anchor $i$, and $z_i$, $z_p$, and $z_a$ denote the anchor, positive, and comparison embeddings, respectively.

The overall objective of DCR\_V is
\begin{equation}
\label{eq:dcrv}
\footnotesize
\mathcal{L}_{\mathrm{DCR\_V}}
=
\mathcal{L}_{\mathrm{CE}}^{(v)}
+
\mathcal{L}_{\mathrm{L2}}^{(v)}
+
\mathcal{L}_{\mathrm{con}}^{(v)}.
\end{equation}

Together, the cosine-based cross-entropy, L2 reconstruction, and contrastive losses improve category discriminability, enforce source-domain prototype alignment, and enhance feature-space separability, respectively, thereby facilitating effective rectification to the detector-compatible source-domain visual distribution.
% 总体而言，基于余弦相似度的交叉熵损失、L2 重构损失和对比损失分别用于增强类别判别性、约束向源域原型对齐以及提升特征空间的可分性，从而共同促进增强原型向与检测器兼容的源域视觉分布有效校正。

\textbf{Text branch: DCR\_T.}
As shown in Fig.~\ref{fig:overview} (b), DCR\_T aligns the augmented visual prototypes with the textual semantic space, keeping the refined features compatible with the frozen text classifier. Given the same input prototype $\hat{\boldsymbol{v}}_{c,j}^{D}$, it produces a refined representation $\tilde{\boldsymbol{v}}_{c,j}^{D,t}$ through a lightweight feed-forward mapping.

DCR\_T is optimized with a cosine-based cross-entropy loss:
\begin{equation}
\label{eq:tce}
\footnotesize
\mathcal{L}_{\mathrm{DCR\_T}}
=
\mathcal{L}_{\mathrm{CE}}^{(t)}
=
-\log
\frac{
\exp\!\left(\cos(\tilde{\boldsymbol{v}}_{c,j}^{D,t}, \boldsymbol{t}_{c})/\tau\right)
}{
\sum\limits_{c'}
\exp\!\left(\cos(\tilde{\boldsymbol{v}}_{c,j}^{D,t}, \boldsymbol{t}_{c'})/\tau\right)
},
\end{equation}
where $\boldsymbol{t}_{c'}$ denotes the text prototypes of all categories, and $\tau$ is the temperature hyperparameter.

Unlike DCR\_V, which focuses on rectifying enhanced prototypes toward a detector-compatible source-domain visual distribution, DCR\_T aims to preserve semantic consistency with the frozen text classifier.
Therefore, we optimize DCR\_T only with a cosine-based cross-entropy loss, which encourages the refined visual prototypes to remain close to the correct text prototypes while being separable from those of other categories.
% 与侧重于将增强原型校正到与检测器兼容的源域视觉分布的 DCR_V 不同，DCR_T 的目标是保持与冻结文本分类器之间的语义一致性。因此，我们仅使用基于余弦相似度的交叉熵损失来优化 DCR_T，该损失鼓励精炼后的视觉原型更接近正确类别的文本原型，同时与其他类别的文本原型保持可分性。

\textbf{Application of DCR.}
As shown in Fig.~\ref{fig:overview} (c), during inference we insert the trained DCR branches after the original classification head. For each proposal feature, DCR\_V and DCR\_T produce branch-specific refined representations, from which visual- and text-prototype-based classification scores are computed, respectively. The final score is obtained by weighted fusion of the two branches. 
In this way, DCR improves compatibility with both the source-domain visual distribution and the frozen text classifier, enabling efficient generalization to novel categories and unseen domains in a \textbf{detector training-free} and \textbf{real-data-free} manner.
% 如图~\ref{fig:overview}(c) 所示，在推理阶段，我们将训练好的 DCR 分支插入到原始分类头之后。对于每个 proposal 特征，DCR_V 和 DCR_T 分别生成各自分支的精炼表示，并据此分别计算基于视觉原型和基于文本原型的分类分数。最终分类分数由这两个分支的分类分数加权融合得到。通过这种方式，DCR 同时提升了特征与源域视觉分布及冻结文本分类器之间的兼容性，从而在detector training-free 的前提下，实现对新类别和未见领域的高效泛化。

%%%%%%%%%%%%%%%%%%%%%%%%%%%%%%%%%%%%%%%%%%%%%%%%%%%%%%%%%%%%%%%%%%%%%%%%%%%%%%%%%%%%%%%%%%%%%%%%%%%%%%%%%%%%%%%%%%%
\subsection{ExRPN for Category and Domain Shift}
\label{sec:RPN}
To alleviate the low recall of novel and domain-shifted objects at the proposal stage—which directly limits subsequent classification and rectification—we adopt the R-RPN design of NoOVD~\cite{zhang2026noovdnovelcategorydiscovery} and extend it to the multi-domain setting, yielding a lightweight inference-time proposal confidence rectification strategy termed ExRPN. It recalibrates proposal confidence before post-processing, increasing the probability that proposals containing novel and domain-shifted objects are preserved for subsequent classification and rectification. As a complement to TGE and DCR, ExRPN provides better support for downstream classification and rectification.

Specifically, based on the K-FPN features used in NoOVD, we extract visual embeddings for proposals from the RPN and obtain proposal representations via RoI Align. We then design a two-stage textual prompting process for coarse domain estimation and foreground confidence reweighting.
% 具体而言，基于 NoOVD 中使用的 K-FPN 特征，我们从 RPN 生成的候选框中提取视觉嵌入，并通过 RoI Align 获得候选区域表征。在此基础上，我们设计了一个两阶段文本提示流程，用于粗粒度领域估计和前景置信度重加权。

\textit{Stage 1: domain estimation.}
We randomly sample a subset of proposal features and compute their similarities to the text embeddings of predefined domain descriptors introduced in TGE (\eg, \textit{a rainy image}, \textit{a foggy image}), pre-extracted by the frozen text encoder. Based on these similarities, we estimate the coarse domain attribute of the input image.

\textit{Stage 2: semantic prompting and foreground ccoring.}
Conditioned on the estimated domain, we use the coarse-grained foreground and background prompts pre-generated by an LLM in NoOVD and compute their cosine similarities with all proposal embeddings. The similarity to the foreground prompt is treated as the foreground confidence, denoted as $\hat{S}_{\textit{Foreg}}$.

We fuse the semantic foreground confidence with $S_{\textit{RPN}}$ to obtain the adaptive proposal confidence:
\begin{equation}
\label{eq:rpn}
\footnotesize
S_{\textit{ExRPN}} = \alpha \cdot S_{\textit{RPN}} + (1 - \alpha) \cdot \hat{S}_{\textit{Foreg}},
\end{equation}
where $\alpha \in [0, 1]$ is a balancing coefficient.

Finally, we replace the original RPN confidence with $S_{\textit{ExRPN}}$ for proposal ranking and filtering. Specifically, proposals are sorted by the adapted confidence in descending order, and the top-$K$ ones (set to 1{,}000, following Faster R-CNN) are retained and fed into the RoI head. In this way, ExRPN improves the probability that proposals containing novel and domain-shifted objects survive post-processing, thereby enhancing proposal recall under cross-category and cross-domain settings with negligible computational overhead, while providing more sufficient effective candidates for subsequent classification and rectification.
%%%%%%%%%%%%%%%%%%%%%%%%%%%%%%%%%%%%%%%%%%%%%%%%%%%%%%%%%%%%%%%%%%%%%%%%%%%%%%%%%%%%%%%%%%%%%%%%%%%%%%%%%%%%%%%%%%%%%%%%%%%%%%%%%%%%%%%%%%%

\subsection{Implementation Details}

We extract visual embeddings of base categories from the RoI classification head of a trained F-ViT detector and build category prototypes. Then, through TGE, we generate augmented visual embeddings covering novel categories and multiple visual domains, with \textbf{500 multi-domain embeddings for each category}. Notably, DCR training requires neither the participation of the detector nor any real data; we train this lightweight DCR module solely with the pre-extracted category prototypes and their extrapolated variants.
% 我们首先从训练好的 F-ViT 检测器的 RoI 分类头中提取基础类别的视觉嵌入，并构建类别原型。随后，通过文本引导外推（Text-Guided Extrapolation, TGE），我们生成覆盖新类别和多种视觉域的增强视觉嵌入（每个类别构造 500 个多域嵌入）。值得注意的是，DCR 的训练既无需 F-ViT 检测器参与，也不依赖真实数据；我们仅利用预先提取的类别原型及其扩展形式来训练这一轻量级 DCR 模块。

The overall objective of DCR consists of the visual rectification branch (DCR\_V) and the semantic rectification branch (DCR\_T):
\begin{equation}
\footnotesize
\mathcal{L}_{\text{total}} = \mathcal{L}_{\text{DCR\_V}} + \mathcal{L}_{\text{DCR\_T}}.
\end{equation} 
For the DCR\_V loss in Eq.~(\ref{eq:dcrv}), we set $\lambda_1 = 1$, $\lambda_2 = 1$, and $\lambda_3 = 1$. For the contrastive loss in Eq.~(\ref{eq:cons}), the temperature coefficient is set to $\tau = 0.7$. For ExRPN in Eq.~(\ref{eq:rpn}), we follow NoOVD~\cite{zhang2026noovdnovelcategorydiscovery} and set the confidence fusion coefficient to $\alpha = 0.5$.

We train DCR on OV-LVIS for 30 epochs on \textbf{a single NVIDIA RTX 3090 GPU} with a batch size of 256. AdamW is used as the optimizer, with an initial learning rate of $10^{-4}$ and a weight decay of 0.1. The total training time is \textbf{only about 30 minutes}.

\begin{figure*}[t!]
	\centering
	\includegraphics[width=1.0\linewidth]{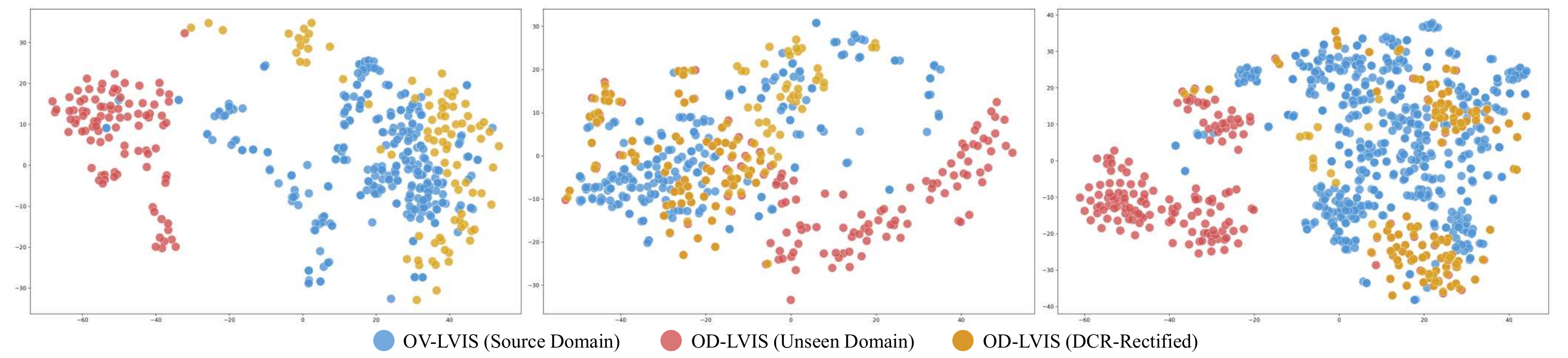}
	\vspace{-20pt}
	\caption{Visualization of OV-LVIS and OD-LVIS object embeddings. After applying DCR, the OD-LVIS embeddings (yellow) are pulled closer to the OV-LVIS source-domain embeddings (blue) than the original OD-LVIS embeddings (red).}
	\label{fig:result_visual}
	\vspace{-10pt}
\end{figure*}

\section{Experiments}
\subsection{Setup}

\textbf{ODOV settings.}
Following ODOVD~\cite{zhang2025odov}, models are trained on a single natural domain (\ie, the OV-LVIS training set) and evaluated on all 15 open domains in OD-LVIS, consistent with the SDGOD setting. For open-vocabulary evaluation, we use the same category split: 405 \textit{frequent} and 461 \textit{common} categories as base categories for training, and 337 \textit{rare} categories as novel categories for testing.

\textbf{Evaluation methods.}
We select several mainstream VLM-based OVD methods and an ODOVD method for comparison on OD-LVIS. Specifically, we include the transfer learning approaches, \ie, F-VLM~\cite{kuo2022f}, OWL-ViT~\cite{minderer2022simple}, F-ViT (CLIPSelf~\cite{wu2023clipself} and DeCLIP~\cite{wang2025declip} as the backbones), MM-OVOD~\cite{xu2023exploring}, and OV-DQUO~\cite{wang2025ov}, and several knowledge distillation methods, \ie, RKDWTF~\cite{bangalath2022bridging}, DK-DETR~\cite{li2023distilling}, RegionCLIP~\cite{zhong2022regionclip}, and region-aware training method YOLO-World~\cite{cheng2024yolo}, YOLOE~\cite{wang2025yoloe}. We also incorporate the ODOV detection method DVtor~\cite{zhang2025odov}.
We also include eight recent DG methods, ALT~\cite{gokhale2023improving}, ABA~\cite{cheng2023adversarial}, NP~\cite{fan2023towards}, MAD~\cite{xu2023multi}, MixStyle~\cite{zhou2024mixstyle}, OA-DG~\cite{lee2024object}, SRA~\cite{xiao2025sample} and PhysAug~\cite{xu2025physaug} for comparison. 

\textbf{Evaluation metrics.}
For evaluation, $AP_f$ and $AP_c$ denote average precision on \textit{frequent} and \textit{common} categories, respectively, while $AP_r$ denotes average precision on \textit{rare} categories. The overall average precision is denoted as $AP$.

\begin{table}[t!]
\setlength{\tabcolsep}{2.5pt}
	\centering
	% \scriptsize
    % \tiny
    \footnotesize
	\caption{Comparison with SOTA on OD-LVIS (\%).}\vspace{-10pt}
	\label{tab:OD-LVIS}
    \begin{threeparttable}
	\begin{tabular}{l|c|c|c|c|c|>{\columncolor{gray!15}}c}
		\hline
		Method    & Backbone  & Training Data  &$AP_f$     &$AP_c$    &$AP_r$  &$AP$  \\\hline
		\multirow{2}*{RegionCLIP~\cite{zhong2022regionclip}} & {RN50}$^*$  & \multirow{2}*{CC3M}  &  16.6      & 13.0      & 9.7     & 13.9 \\
		  & {RN50$\times$4}$^*$ &  & 19.5       & 15.8      & 12.4     & 16.7 \\
		\hline
		\multirow{2}*{OWL-ViT~\cite{minderer2022simple}} & ViT-B/16  & \multirow{2}*{O365 + VG}  & 13.1       & 13.9      & 13.2     & 13.5\\
		  & ViT-L/14 &   & 22.1      & 21.6      & 19.9     & 21.5  \\
		\hline
		\multirow{4}*{RKDWTF~\cite{bangalath2022bridging}} & {RN50}$^*$ Base  & \multirow{4}*{LVIS-base + IN-L}  & 14.6  & 12.4  & 8.7  & 12.6 \\
		  & {RN50}$^*$ RKDPIS  &  & 13.4 & 12.1 & 10.3 & 12.3 \\
		  & {RN50}$^*$ WTF &  &  14.0  & 12.5 & 11.3 & 12.9 \\
		  & {RN50}$^*$ WTF8x  &  & 15.8 & 14.3 & 11.9 & 14.5 \\
		\hline
        DK-DETR~\cite{li2023distilling} & RN50 & LVIS-all  & 21.1 &  19.4 & 15.3 & 19.4 \\
        \hline 
		  \multirow{2}*{MM-OVOD~\cite{xu2023exploring}} & \multirow{2}*{{RN50}$^*$ (Agg)}  & LVIS-base & 20.5 & 19.8  & 14.0 & 19.0 \\
        \cline{3-7}
		   &  & LVIS-base + IN-L  & 20.4  & 20.4  & 15.9  & 19.6\\
		\hline
        YOLO-World~\cite{cheng2024yolo} & YOLOv8-L$^*$ & \multirow{2}*{O365 + GoldG}  &21.9 &19.1 &19.3 &20.2 \\
        \cline{1-2}\cline{4-7}
        YOLOE~\cite{wang2025yoloe} & YOLOv11-L$^*$   &  &13.7  &8.6  &6.8  &10.3  \\
		\hline
        \multirow{2}*{OV-DQUO~\cite{wang2025ov}} & ViT-B/16 & \multirow{17}*{LVIS-base} &12.8  &14.8  &14.8  &14.0  \\
        % \cline{4-7}
        & ViT-L/14 & &16.4  &20.6  &21.2  &19.1  \\ 
		\cline{1-2}\cline{4-7}
		  F-VLM (CLIP)~\cite{kuo2022f}  & \multirow{3}*{RN50$\times$16} & & 16.7 & 14.4 & 13.7 & 15.2  \\
        DVtor (CLIP)~\cite{zhang2025odov} & & &17.6  &16.9 &15.8 &17.0  \\ 
        \textbf{ExDet (CLIP)} & & &\textbf{17.8}&\textbf{18.8}&\textbf{19.0}&\textbf{18.4}  \\ 
		\cline{1-2}\cline{4-7}
		F-ViT (CLIPSelf)~\cite{wu2023clipself} & \multirow{3}*{ViT-B/16} & & 17.1 & 12.0  & 12.2 &14.0  \\
        DVtor (CLIPSelf)~\cite{zhang2025odov} &  &  &19.0  &14.3 &14.0 &16.1  \\ 
        \textbf{ExDet (CLIPSelf)} & & &\textbf{19.2}&\textbf{18.0}&\textbf{16.1} &\textbf{18.1}  \\ 
        \cline{1-2}\cline{4-7}
		  F-ViT (CLIPSelf)~\cite{wu2023clipself} & \multirow{3}*{ViT-L/14} & &22.5   & 21.3 & 20.2 & 21.6 \\ 
        DVtor (CLIPSelf)~\cite{zhang2025odov} &  &  &\textbf{23.9}  &22.9 &21.6 &23.1  \\
        \textbf{ExDet (CLIPSelf)} & & &\textbf{23.9}&\textbf{24.3}&\textbf{25.9}&\textbf{24.5}  \\ 
        \cline{1-2}\cline{4-7}
		F-ViT (DeCLIP)~\cite{wang2025declip} & \multirow{3}*{ViT-B/16}  &  &17.8  &12.9  &13.2  &14.9  \\
        DVtor (DeCLIP)~\cite{zhang2025odov}   &  &  &\textbf{19.5}  &14.7  &15.5  &16.7  \\
        \textbf{ExDet (DeCLIP)} & & &19.4&\textbf{18.9}&\textbf{17.3}&\textbf{18.8}  \\
        \cline{1-2}\cline{4-7}
		  F-ViT (DeCLIP)~\cite{wang2025declip} & \multirow{3}*{ViT-L/14} & &23.0  &21.7  &21.4  &22.2  \\
        DVtor (DeCLIP)~\cite{zhang2025odov} &  &  &24.9  &22.6  &23.2  &23.6  \\ 
        \textbf{ExDet (DeCLIP)} & & &\textbf{25.3}&\textbf{24.7}&\textbf{26.3}&\textbf{25.2}  \\
		\hline
	\end{tabular} 
        \begin{tablenotes}
            \footnotesize
            % \scriptsize
            % \tiny
            \item Notes: IN-L denotes the inclusion of images corresponding to the 997 categories shared between ImageNet-21k-P~\cite{ridnik2021imagenet} and LVIS~\cite{gupta2019lvis}, `$^*$' indicates that the backbone is not initialized with CLIP, O365 is an abbreviation for Objects365~\cite{shao2019objects365}, CC3M~\cite{sharma2018conceptual}, GoldG~\cite{kamath2021mdetr}, and VG~\cite{krishna2017visual} are all publicly available datasets.  
        \end{tablenotes}
    \end{threeparttable} \vspace{-7pt}
\end{table}

\begin{table}[t!] 
	\centering
    \footnotesize
	% \scriptsize
    % \tiny
    \caption{Comparison of DG methods on OD-LVIS (\%).}  \vspace{-10pt}
	\label{tab:dg}
	\begin{tabular}{l|c|c|c|>{\columncolor{gray!15}}c}
		\hline
		Method      &$AP_f$     &$AP_c$    &$AP_r$    &$AP$  \\\hline
        F-ViT (DeCLIP) (ViT-B/16)                  &17.8  &12.9  &13.2  &14.9 \\
        ~~~ + ALT~\cite{gokhale2023improving}     &18.2  &13.2  &13.1  &15.1 \\
        ~~~ + ABA~\cite{cheng2023adversarial}     &18.0  &13.4  &13.8  &15.3 \\
        ~~~ + NP~\cite{fan2023towards}            &18.6  &14.0  &14.2  &15.8 \\
        ~~~ + MAD~\cite{xu2023multi}              &18.8  &13.9  &14.1  &15.9 \\
        ~~~ + MixStyle~\cite{zhou2024mixstyle}    &17.9  &13.3  &13.5  &15.1 \\
        ~~~ + OA-DG~\cite{lee2024object}          &17.2  &14.5  &13.2  &15.3 \\
        ~~~ + SRA~\cite{xiao2025sample}           &16.9  &14.2  &14.0  &15.2 \\
        ~~~ + PhysAug~\cite{xu2025physaug}        &18.3  &13.6  &13.7  &15.5 \\ \hline
        DVtor~\cite{zhang2025odov}                &\textbf{19.5}  &14.7  &15.5  &16.7 \\ \hline
        \textbf{ExDet}                            &19.4  &\textbf{18.9}  &\textbf{17.3}  &\textbf{18.8}  \\
		\hline
	\end{tabular} 
 \vspace{-18pt}
\end{table}

\begin{table}[t!]
\setlength{\tabcolsep}{2.5pt} 
	\centering
    \footnotesize
	% \scriptsize
    % \tiny
	\caption{Comparative results on OV-LVIS (\%).} \vspace{-10pt}
	\label{tab:LVIS}
    \begin{tabular}{l|c|c|c|c|c|>{\columncolor{gray!15}}c}
        \hline
        Method    & Backbone & Training Data  &$AP_f$     &$AP_c$    &$AP_r$  &$AP$  \\\hline
        \multirow{2}*{RegionCLIP~\cite{zhong2022regionclip}} & {RN50}$^*$ &\multirow{2}*{CC3M} & 34.0 & 27.4 & 17.1 &  28.2 \\
          & {RN50$\times$4}$^*$ & & 36.9 & 32.1 & 22.0 & 32.3 \\
        \hline
        \multirow{2}*{OWL-ViT~\cite{minderer2022simple}} & ViT-B/16 & \multirow{2}*{O365 + VG}  & - & - & 20.6 &  27.2 \\
          & ViT-L/14 &  & - & - & 31.2 & 34.6 \\
        \hline
        \multirow{4}*{RKDWTF~\cite{bangalath2022bridging}} & {RN50}$^*$ Base & \multirow{4}*{LVIS-base + IN-L} & 26.4 & 19.4 & 12.2 & 20.9 \\
          & {RN50}$^*$ RKDPIS &  & 25.5 & 20.9 & 17.3 & 22.1 \\
          & {RN50}$^*$ WTF &  & 26.7 & 21.4 & 17.1 & 22.8 \\
          & {RN50}$^*$ WTF8x  &  & 29.1 & 25.0 & 21.1 & 25.9 \\
        \hline
        \multirow{2}*{MM-OVOD~\cite{xu2023exploring}} 
        % & {RN50}$^*$(Avg)  & \multirow{2}*{LVIS-base}  & - & - & 20.7 & 30.5 \\
          & \multirow{2}*{{RN50}$^*$(Agg)}  &LVIS-base  & - & - & 19.3 & 30.6 \\
          \cline{3-7}
          % & {RN50}$^*$(Avg)  & \multirow{2}*{LVIS-base + IN-L} & - & - & 26.5 & 32.8 \\
          &   & LVIS-base + IN-L & - & - & 27.3 & 33.1 \\
        \hline
        DK-DETR~\cite{li2023distilling} & RN50  & LVIS-all & 40.2 & 32.0 & 22.2 & 33.5 \\
        \hline
        YOLO-World~\cite{cheng2024yolo} & {YOLOv8-L}$^*$ & \multirow{2}*{O365 + GoldG}  &35.4 &24.9 &22.9 &28.7 \\
        \cline{1-2}\cline{4-7}
        YOLOE~\cite{wang2025yoloe} & {YOLOv11-L}$^*$ & &36.5 &35.0 &29.1 &35.2 \\
        \hline 
        \multirow{2}*{OV-DQUO~\cite{wang2025ov}} & ViT-B/16 & \multirow{17}*{\makecell{LVIS-base}} &23.8  &27.7   &29.4  &26.5  \\
          & ViT-L/14 & &28.5  &36.0  &39.5  &33.7  \\ 
        \cline{1-2}\cline{4-7}
        F-VLM (CLIP)~\cite{kuo2022f}  & \multirow{3}*{RN50$\times$16} &  & - & - & 30.4 & 32.1 \\
        DVtor (CLIP)~\cite{zhang2025odov}  &  &  &33.0  & 34.1 & 33.1 & 33.5  \\ 
        \textbf{ExDet (CLIP)} &  &  &\textbf{33.5} &\textbf{35.6}  &\textbf{35.0}  &\textbf{34.7}   \\ 
        \cline{1-2}\cline{4-7}
        F-ViT (CLIPSelf)~\cite{wu2023clipself} & \multirow{3}*{ViT-B/16}  &  &29.1  &21.8 &25.3 &25.2  \\
        DVtor (CLIPSelf)~\cite{zhang2025odov}  &  &  &30.4 &23.2 &26.3 &26.6  \\ 
        \textbf{ExDet (CLIPSelf)} &  &  &\textbf{31.1} &\textbf{26.3}  &\textbf{30.2}  &\textbf{28.8}   \\ 
        \cline{1-2}\cline{4-7}
        F-ViT (CLIPSelf)~\cite{wu2023clipself} & \multirow{3}*{ViT-L/14} & &35.6  &34.6  &34.9  &35.1  \\
        DVtor (CLIPSelf)~\cite{zhang2025odov}  &  & &36.9  &35.8  &36.4  &36.3  \\ 
        \textbf{ExDet (CLIPSelf)} &  &  &\textbf{38.1} &\textbf{37.0}  &\textbf{39.7}  &\textbf{37.9}   \\ 
        \cline{1-2}\cline{4-7}
        F-ViT (DeCLIP)~\cite{wang2025declip} & \multirow{3}*{ViT-B/16} & & 29.8  &22.4  & 26.8 & 26.0 \\
        DVtor (DeCLIP)~\cite{zhang2025odov}  &  & &31.0 &23.7 &28.1 &27.3   \\ 
        \textbf{ExDet (DeCLIP)} &  &  &\textbf{31.6} &\textbf{27.1}  &\textbf{30.5}  &\textbf{29.4}   \\ 
        \cline{1-2}\cline{4-7}
        F-ViT (DeCLIP)~\cite{wang2025declip} & \multirow{3}*{ViT-L/14} & & 36.5  & 35.2  & 37.2  &36.0\\ 
        DVtor (DeCLIP)~\cite{zhang2025odov}   &  & &37.6 &35.9 &39.0 &37.1  \\ 
        \textbf{ExDet (DeCLIP)} &  &  &\textbf{38.2}  &\textbf{37.0}  &\textbf{40.1}  &\textbf{38.0}   \\ 
        \hline  
    \end{tabular} \vspace{-15pt}
\end{table}

\subsection{Main Results on OD-LVIS}
\textbf{Comparison with ODOVD and OVD methods on OD-LVIS.}
Table~\ref{tab:OD-LVIS} compares our method with ODOVD and OVD methods on OD-LVIS. 
Our method achieves the best overall performance across multiple detector instantiations. 
Built upon F-ViT (DeCLIP ViT-L/14, 304.43M), ExDet attains the best results among all compared methods, reaching 25.3\%, 24.7\%, 26.3\%, and 25.2\% on the \textit{frequent}, \textit{common}, \textit{rare}, and overall categories, respectively. Compared with the corresponding F-ViT (DeCLIP) baseline, this yields gains of 2.3\%, 3.0\%, 4.9\%, and 3.0\%; compared with DVtor, the gains remain 0.4\%, 2.1\%, 3.1\%, and 1.6\%, respectively.
Our method also generalizes well to smaller backbones. 
With F-ViT (DeCLIP ViT-B/16, 86.26M), ExDet improves the baseline by 1.6\%, 6.0\%, 4.1\%, and 3.9\% on \textit{frequent}, \textit{common}, \textit{rare}, and overall $AP$, respectively, and surpasses DVtor by 4.2\%, 1.8\%, and 2.1\% on \textit{common}, \textit{rare}, and overall $AP$. 
When integrated into F-VLM (CLIP RN50$\times$16, 167.33M), our method improves the baseline by 1.1\%, 4.4\%, 5.3\%, and 3.2\% on \textit{frequent}, \textit{common}, \textit{rare}, and overall $AP$, respectively, and outperforms DVtor by 1.9\%, 3.2\%, and 1.4\% on \textit{common}, \textit{rare}, and overall $AP$.
Across different detector variants, ExDet consistently outperforms the corresponding baselines and DVtor, verifying the strong transferability of DCR and ExRPN, as well as their effectiveness in improving open-domain open-vocabulary generalization.

\begin{table}[t!]
	\centering
    \footnotesize
    % \tiny
	\caption{Cross-dataset results on Objects365 (\%).} \vspace{-10pt}
	\label{tab:O365}
	\begin{tabular}{l|c|c|c|>{\columncolor{gray!15}}c}
        \hline
        Method  &Backbone     & Training Data   &$AP_r$    &$AP$  \\\hline
        \multirow{2}*{Detic~\cite{zhou2022detecting}}   & \multirow{4}*{{RN50}$^*$}  & LVIS-all    & 9.5   & 13.9    \\
        &  & LVIS-all + IN-L   &12.4    & 15.6    \\
        \cline{1-1}\cline{3-5}
        \multirow{2}*{MM-OVOD~\cite{xu2023exploring}} &   & LVIS-all   & 10.1 & 14.8     \\
         &  & LVIS-all + IN-L   & 13.1 & 16.6    \\
        \hline
        F-VLM (CLIP)~\cite{kuo2022f} & \multirow{3}*{RN50$\times$16}   &\multirow{15}*{LVIS-base}  & 14.9   &16.2     \\
        DVtor (CLIP)~\cite{zhang2025odov} &    &  &16.2  &17.9     \\
        \textbf{ExDet (CLIP)} &   &  &\textbf{18.6}  &\textbf{20.3}    \\
        \cline{1-2}\cline{4-5}
	    F-ViT (CLIPSelf)~\cite{wu2023clipself} & \multirow{3}*{ViT-B/16}    &    &16.8  & 19.0     \\
        DVtor (CLIPSelf)~\cite{zhang2025odov} &   &  &17.5  &19.8    \\
        \textbf{ExDet (CLIPSelf)} &   &  &\textbf{18.4} &\textbf{20.5}     \\
        \cline{1-2}\cline{4-5}
	    F-ViT (CLIPSelf)~\cite{wu2023clipself}  & \multirow{3}*{ViT-L/14}   &   &21.7  &23.7   \\
        DVtor (CLIPSelf)~\cite{zhang2025odov} &   &  &22.3  &24.0     \\
        \textbf{ExDet (CLIPSelf)} &   &  & \textbf{23.8} &\textbf{25.1}    \\
        \cline{1-2}\cline{4-5}
        F-ViT (DeCLIP)~\cite{wang2025declip} & \multirow{3}*{ViT-B/16}  &   & 17.6  & 20.2   \\
        DVtor (DeCLIP)~\cite{zhang2025odov} &   &  &18.7 &21.1    \\
        \textbf{ExDet (DeCLIP)} &   &  &\textbf{19.8}  &\textbf{22.6}    \\
        \cline{1-2}\cline{4-5}
	    F-ViT (DeCLIP)~\cite{wang2025declip} & \multirow{3}*{ViT-L/14}   &   & 22.3  & 24.5   \\
        DVtor (DeCLIP)~\cite{zhang2025odov} &   &  &23.7  &25.0    \\
        \textbf{ExDet (DeCLIP)} &   &  &\textbf{25.1}  &\textbf{26.2}    \\
	\hline 
	\end{tabular} 
 \vspace{-5pt}
\end{table}

\begin{table}[t!]
    \centering
    % \tiny
    \footnotesize
    \caption{Cross-dataset results on MSOSB (\%).}\vspace{-10pt}
    \label{tab:msosb}
	\begin{tabular}{l|c|c|c|>{\columncolor{gray!15}}c}
        \hline
        Method  &Backbone  & Training Data   &$AP_r$    &$AP$  \\\hline
        F-VLM (CLIP)~\cite{kuo2022f} & \multirow{3}*{RN50$\times$16}  &\multirow{15}*{LVIS-base} &31.3    &37.2      \\
        DVtor (CLIP)~\cite{zhang2025odov} &  &  & 32.8  & 39.0   \\
        \textbf{ExDet (CLIP)} &  &  &\textbf{33.6}   &\textbf{39.7}  \\
        \cline{1-2}\cline{4-5}
	    F-ViT (CLIPSelf)~\cite{wu2023clipself} & \multirow{3}*{ViT-B/16}  &  &29.8  & 35.6    \\
        DVtor (CLIPSelf)~\cite{zhang2025odov} &  &  &31.4  &37.2      \\
        \textbf{ExDet (CLIPSelf)} &  &  &\textbf{32.6}   &\textbf{37.9}  \\
        \cline{1-2}\cline{4-5}
	    F-ViT (CLIPSelf)~\cite{wu2023clipself} & \multirow{3}*{ViT-L/14}  &  &38.2  &42.6    \\
        DVtor (CLIPSelf)~\cite{zhang2025odov} &  &  &39.8  &43.9    \\
        \textbf{ExDet (CLIPSelf)} &  &  &\textbf{40.9}   &\textbf{44.3}  \\
        \cline{1-2}\cline{4-5}
        F-ViT (DeCLIP)~\cite{wang2025declip} & \multirow{3}*{ViT-B/16} &  & 30.8  & 36.3    \\
        DVtor (DeCLIP)~\cite{zhang2025odov} &  &  &32.5   &37.9    \\
        \textbf{ExDet (DeCLIP)} &  &  &\textbf{34.1}   &\textbf{38.8}  \\
        \cline{1-2}\cline{4-5}
	    F-ViT (DeCLIP)~\cite{wang2025declip}& \multirow{3}*{ViT-L/14}  &  &38.8   & 43.3   \\
        DVtor (DeCLIP)~\cite{zhang2025odov} &  &  &40.1  & 45.0    \\
        \textbf{ExDet (DeCLIP)} &  &  &\textbf{41.5}   &\textbf{45.7}  \\
        \hline
    \end{tabular} \vspace{-20pt}
\end{table}

As shown in Fig.~\ref{fig:result_visual}, we visualize object embeddings of three representative categories from OV-LVIS, OD-LVIS, and DCR-rectified OD-LVIS. Compared with the original OD-LVIS embeddings, the DCR-rectified embeddings are clearly closer to the OV-LVIS source-domain clusters. This shows that DCR effectively narrows the visual embedding distribution gap and improves feature compatibility for cross-domain classification.
% 如图~\ref{fig:result_visual} 所示，我们可视化了三个代表性类别在 OV-LVIS、OD-LVIS 及经过 DCR 校正后的 OD-LVIS 上的目标嵌入。相比原始 OD-LVIS 嵌入，经过 DCR 校正后的嵌入明显更接近 OV-LVIS 的源域聚类中心。这表明 DCR 能够有效缩小视觉嵌入的分布差距，并提升特征对跨域目标的分类兼容性。

\textbf{Comparison with DG methods on OD-LVIS}
Table~\ref{tab:dg} compares our method with F-ViT and eight DG methods on OD-LVIS under the same DeCLIP ViT-B/16 backbone. ExDet achieves the best overall performance, obtaining 19.4\%, 18.9\%, 17.3\%, and 18.8\% on the \textit{frequent}, \textit{common}, \textit{rare}, and overall categories, respectively. Compared with F-ViT and DVtor, it improves the overall $AP$ by 3.9 and 2.1 points, respectively, with especially notable gains on the \textit{common} and \textit{rare} categories. This verifies the strong transferability and generalization ability of our method under domain shifts.

\subsection{Additional Evaluation on OV-LVIS}
Table~\ref{tab:LVIS} reports the comparison on the OV-LVIS validation set. 
ExDet consistently achieves the best overall performance across multiple detector instantiations. 
In particular, built upon F-ViT (DeCLIP ViT-L/14), it delivers the best results among all compared methods, reaching 38.2\%, 37.0\%, 40.1\%, and 38.0\% on \textit{frequent}, \textit{common}, \textit{rare}, and overall categories, respectively. Relative to the corresponding F-ViT (DeCLIP) baseline, this yields gains of 1.7\%, 1.8\%, 2.9\%, and 1.4\%; relative to DVtor, the gains remain 0.6\%, 1.1\%, 1.1\%, and 0.9\%, respectively.
ExDet also generalizes well to smaller backbones. 
With F-ViT (DeCLIP ViT-B/16), it improves the baseline by 1.8\%, 2.9\%, 3.7\%, and 3.4\% on $AP_f$, $AP_c$, $AP_r$, and overall $AP$, respectively, and surpasses DVtor by 0.6\%, 3.4\%, 2.4\%, and 2.1\%. 
With F-ViT (CLIPSelf ViT-L/14), ExDet achieves 38.1\%, 37.0\%, 39.7\%, and 37.9\%, improving the baseline by 2.5\%, 2.4\%, 4.8\%, and 2.8\%, and DVtor by 1.2\%, 1.2\%, 3.3\%, and 1.6\%, respectively. 
Even under the RN50$\times$16 setting, it improves F-VLM by 0.5\%, 2.5\%, 4.6\%, and 1.5\% on $AP_f$, $AP_c$, $AP_r$, and overall $AP$, respectively, while still outperforming DVtor by 0.5\%, 1.5\%, 1.9\%, and 1.2\%.
Although the domain gap in OV-LVIS is less pronounced than that in OD-LVIS, ExDet still brings consistent gains across different detectors and backbones. This indicates that the proposed framework not only improves robustness to domain variations, but also enhances novel-category generalization by leveraging category- and domain-aware proxy visual prototypes together with Detector-Compatible Rectification.

\subsection{Cross-dataset Generalization Results}
\textbf{Main results on Objects365.}
Table~\ref{tab:O365} reports the cross-dataset transfer results from OV-LVIS to Objects365, where all models are directly evaluated on the validation set without adaptation. Following MM-OVOD, we define the bottom one-third categories by frequency as \textit{rare} categories. Detic and MM-OVOD are trained on LVIS-all, with the In-L variants further using ImageNet-21k-P~\cite{ridnik2021imagenet}, while F-ViT, DVtor, and ExDet are trained only on the OV-LVIS base categories. ExDet achieves the best transfer performance across different backbones and VLM initializations. With F-ViT (DeCLIP ViT-L/14), it reaches 25.1\% $AP_r$ and 26.2\% $AP$, surpassing the baseline by 2.8\% and 1.7\%, and DVtor by 1.4\% and 1.2\%, respectively. Similar gains are observed under CLIP and CLIPSelf settings. Although smaller than on OD-LVIS and OV-LVIS, these improvements remain consistent, confirming better cross-dataset robustness and novel-category generalization.

\begin{table}[t!]\vspace{-0pt}
    \centering
    % \small
    \footnotesize
    % \scriptsize
    % \tiny
    \caption{Ablation study results on OD-LVIS (\%).} \vspace{-10pt}
    \label{tab:ablation}
    \begin{tabular}{l|c|c|c|>{\columncolor{gray!15}}c}
        \hline
        \textbf{Method} &$AP_f$     &$AP_c$    &$AP_r$    &$AP$ \\
        \hline
        F-ViT (DeCLIP) (ViT-B/16)   &17.8&12.9&13.2&14.9 \\
        \hspace{1em}+ DCR\_V          &18.2&15.8&14.2&16.5 \\
        \hspace{1em}+ DCR\_T           &18.6&14.7&16.2&16.5 \\
        \hspace{1em}+ DCR\_V + DCR\_T   &18.8&16.9&16.7&17.6 \\
        \hspace{1em}+ ExRPN        &18.2&14.3&15.6 &16.0\\
        \textbf{ExDet} (DCR\_V + DCR\_T + ExRPN)   &19.4&18.9&17.3&18.8 \\
        \hline
    \end{tabular} \vspace{-15pt}
\end{table}

\begin{figure*}[t!]
	\centering
	\includegraphics[width=0.97\linewidth]{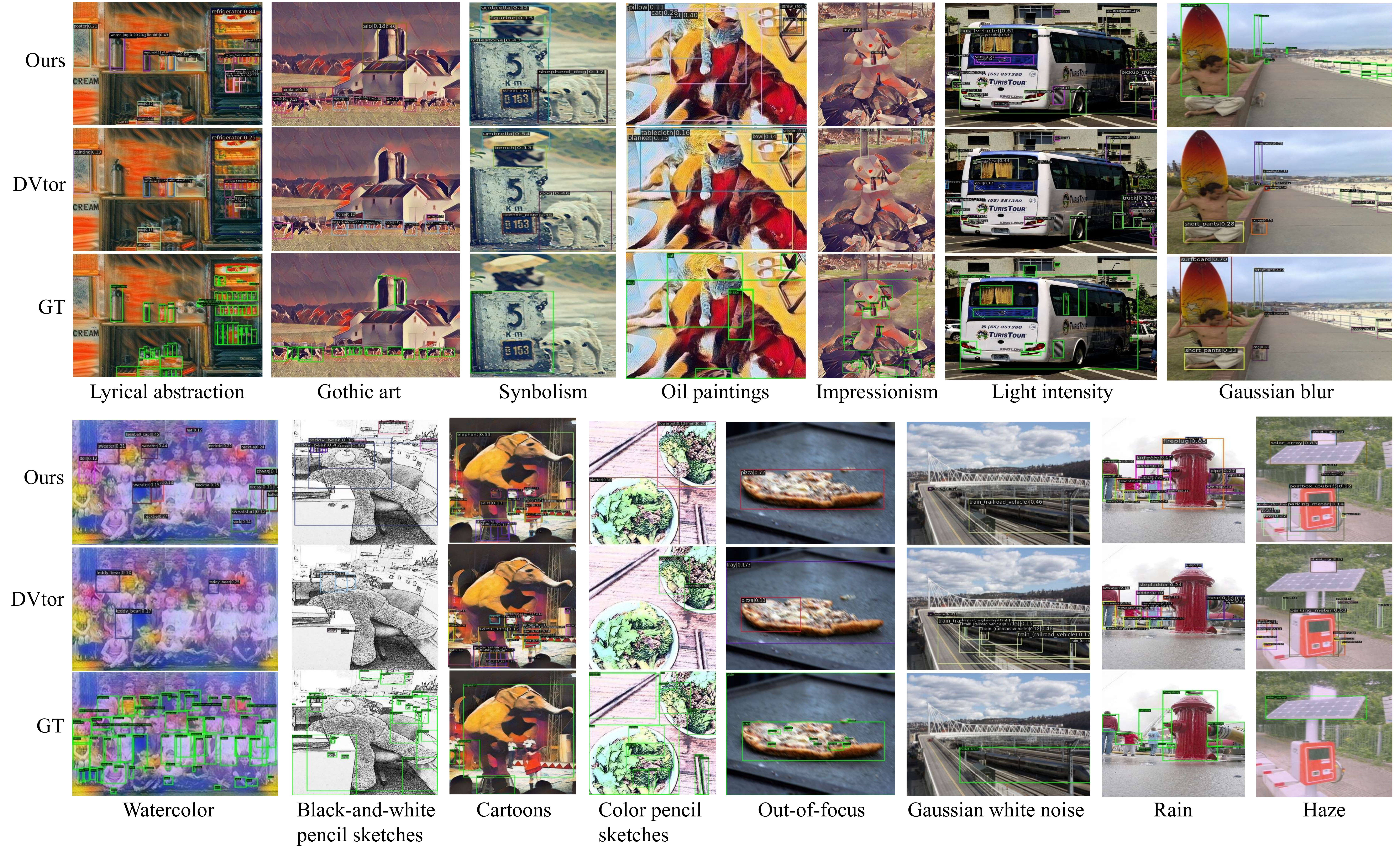}\vspace{-10pt}
	\caption{Qualitative comparison of our method, DVtor, and GT under diverse domain shifts and challenging imaging conditions.}
	\label{fig:results}
    \vspace{-10pt}
\end{figure*}

\textbf{Main results on MSOSB.}
MSOSB is a multi-style benchmark with 5 visual styles and the same 80 categories as COCO. Following category frequency, we treat the 20 less frequent categories as \textit{rare} and the remaining 60 as regular. Table~\ref{tab:msosb} reports the corresponding cross-dataset transfer results, where all models are directly tested without finetuning. ExDet achieves the best performance across all settings. With F-ViT (DeCLIP ViT-L/14), it reaches 41.5\% $AP_r$ and 45.7\% $AP$, surpassing the baseline by 2.7\% and 2.4\%, and DVtor by 1.4\% and 0.7\%, respectively. Similar gains are observed with CLIP and CLIPSelf, confirming strong generalization to unseen datasets and visual styles without target-dataset adaptation.

\subsection{Ablation Study}
\textbf{Component analysis.} Table~\ref{tab:ablation} reports the ablation results for F-ViT (DeCLIP ViT-B/16) on OD-LVIS to evaluate the contributions of DCR\_V, DCR\_T, and ExRPN. Each component improves the baseline. DCR\_V raises $AP$ from 14.9\% to 16.5\%, while DCR\_T achieves the same gain and improves rare categories (16.2\% vs.\ 14.2\%). Combining DCR\_V and DCR\_T further improves $AP$ to 17.6\%, showing their complementarity for cross-domain novel-category detection. ExRPN alone improves the baseline to 16.0\% $AP$, with clear gains on both common and rare categories, showing its effectiveness in improving proposal quality under domain shift. The full model performs best, reaching 19.4\%, 18.9\%, 17.3\%, and 18.8\% on $AP_f$, $AP_c$, $AP_r$, and $AP$, respectively. This confirms that DCR and ExRPN are complementary: the former improves category- and domain-aware representation alignment, while the latter strengthens localization and recall in unseen domains.
% \textbf{组件分析.}
% 表~\ref{tab:ablation} 给出了基于 F-ViT (DeCLIP) with ViT-B/16 在 OD-LVIS 上的消融实验结果，用于评估 DCR$_V$、DCR$_T$ 和 ExRPN 的贡献。
% 从结果可以看出，每个组件单独引入时都能有效提升基线检测器的性能。具体而言，引入 DCR$_V$ 后，整体 AP 从 14.9\% 提升至 16.5\%；而 DCR$_T$ 在取得相同整体增益的同时，在 rare 类别上带来了更明显的提升（16.2\% vs.\ 14.2\%）。当 DCR$_V$ 与 DCR$_T$ 结合使用时，整体性能进一步提升至 17.6\% AP，表明视觉侧与文本侧的重对齐在跨域新类别检测中具有良好的互补性。
% 单独引入 ExRPN 也可将基线提升至 16.0\% AP，并在 common 和 rare 类别上均带来明显增益，验证了其在域偏移条件下提升候选框质量的有效性。
% 最终，同时结合 DCR$_V$、DCR$_T$ 和 ExRPN 时取得了最佳性能，在 $AP_f$、$AP_c$、$AP_r$ 和 AP 上分别达到 19.4\%、18.9\%、17.3\% 和 18.8\%。这进一步说明 DCR 与 ExRPN 具有较强的互补性：前者主要改善类别感知与域感知的表征对齐，后者则进一步增强了未见域中的定位能力与召回能力。

\textbf{Inference efficiency analysis.}
We evaluate speed on a single RTX 3090 GPU. The baseline F-ViT (DeCLIP ViT-B/16) runs at 21.18 FPS, while the full model with DCR and ExRPN achieves 12.55 FPS. Despite the extra overhead, it improves overall $AP$ by 3.9\%, showing a reasonable accuracy--efficiency trade-off. Under the same setting, MM-OVOD and OV-DQUO run at 2.09 FPS and 2.31 FPS, respectively, while DVtor achieves 15.17 FPS. Although our method is moderately slower than DVtor, it delivers higher accuracy and remains substantially more efficient than other methods.
% \textbf{推理效率分析.}
% 我们在单张 NVIDIA RTX 3090 GPU 上评估了模型的推理速度。基线模型 F-ViT (DeCLIP) with ViT-B/16 的推理速度为 21.18 FPS，而加入 DCR 和 ExRPN 后的完整模型推理速度为 12.55 FPS。尽管所引入的模块带来了一定的额外计算开销，但同时也带来了 3.9\% 的整体 $AP$ 提升，体现了较为合理的精度--效率权衡。
% 我们还在相同设置下与若干代表性方法进行了运行速度对比。MM-OVOD 和 OV-DQUO 的推理速度分别仅为 2.09 FPS 和 2.31 FPS，而 DVtor 为 15.17 FPS。尽管我们的方法相比 DVtor 略慢，但能够取得更高的检测精度，并且相较于其他现有方法仍保持了显著更优的运行效率。

\setlength{\columnsep}{6pt}
\begin{wraptable}{r}{0.13\textwidth}\vspace{-13pt}
    \centering
    \footnotesize
    \setlength{\tabcolsep}{2.1pt}
    \caption{Recall on OD-LVIS (\%).}\vspace{-10pt}
    \label{tab:AR}
    \begin{tabular}{l|c|c}
        \hline
        Method & $AR_r$ & $AR$\\\hline
        RPN            & 53.9 & 56.2 \\
        R-RPN          & 58.6 & 61.1 \\
        \textbf{ExRPN} & \textbf{63.2}  & \textbf{65.5} \\
        \hline
    \end{tabular}
    \vspace{-10pt}
\end{wraptable}
% 为量化评估 ExRPN 对开放域目标的覆盖能力，我们采用 \textit{Recall}@IoU$\geq$0.5 作为评价指标。如表~\ref{tab:AR} 所示，在 OD-LVIS 上, 基于F-ViT (DeCLIP ViT-B/16)，Ex-RPN 在rare类和整体召回率上均显著优 于R-RPN 和原始 RPN。
% \textbf{Advantages of ExRPN.} To evaluate the coverage of ExRPN on open-domain open-vocabulary objects, we use \textit{Recall}@IoU$\geq$0.5 as the evaluation metric. As shown in Table~\ref{tab:AR}, built upon F-ViT (DeCLIP ViT-B/16), ExRPN achieves significantly higher recall than both R-RPN and the original RPN on OD-LVIS.
\textbf{Advantages of ExRPN.} To evaluate ExRPN's coverage of open-domain open-vocabulary objects, we use \textit{Recall}@IoU$\geq$0.5. As shown in Table~\ref{tab:AR}, on OD-LVIS with F-ViT (DeCLIP ViT-B/16), Ex-RPN consistently outperforms both R-RPN and the original RPN in rare-category recall $AR_r$ and overall recall $AR$.

\section{Visualized Results}
In Fig.~\ref{fig:results}, we visualize the detection results of GT, DVtor, and our method under diverse domain shifts and imaging degradations. Compared with DVtor, our method detects more objects with more accurate localization, showing stronger robustness and generalization in ODOVD. Under style shifts such as lyrical abstraction, gothic art, symbolism, oil paintings, watercolor, and cartoons, DVtor often misses meaningful objects or produces incomplete detections, whereas our method identifies more relevant objects. Similar gains are observed under challenging conditions, including Gaussian blur, Gaussian white noise, rain, haze, and out-of-focus degradation, where our method preserves better recall and localization quality. These results suggest that ExDet is more robust to both domain variation and image corruption. Nevertheless, compared with GT, it still misses some objects in highly challenging scenarios, indicating that ODOVD remains difficult and that object coverage and recognition accuracy still have room for improvement.

% 如图~\ref{fig:results} 所示，我们展示了 ground-truth（GT）、DVtor 以及本文方法在多种域偏移与成像退化条件下的检测可视化结果。相比 DVtor，我们的方法在大多数情况下能够检测出更多目标，并取得更准确的定位结果，体现出更强的 ODOV 检测鲁棒性与泛化能力。
% 具体而言，在 lyrical abstraction、gothic art、symbolism、oil paintings、watercolor 和 cartoons 等风格迁移场景下，DVtor 往往会遗漏具有明确语义的目标，或者仅产生不完整的检测结果，而我们的方法能够识别出更多相关目标。在 Gaussian blur、Gaussian white noise、rain、haze 以及 out-of-focus 等具有挑战性的成像退化条件下，也能观察到类似现象：我们的方法在目标召回率和定位质量方面整体表现更优。
% 这些结果表明，得益于所提出的类别感知与域感知代理视觉原型，以及分布兼容的重对齐机制，ExDet 对域变化和图像退化都具备更好的适应能力。
% 尽管如此，与 GT 标注相比，我们的方法在高难场景下仍然存在一定的漏检现象，这说明 ODOV 检测仍然是一个具有较大挑战性的问题，在目标覆盖率和识别准确性方面仍有进一步提升空间。

\section{Conclusion}
We propose ExDet, a lightweight framework for ODOVD. Built upon a frozen CLIP-based two-stage detector, ExDet consists of Text-Guided Extrapolation (TGE), Detector-Compatible Rectification (DCR), and ExRPN. 
TGE infers category- and domain-aware proxy visual prototypes from text. DCR is learned independently under the supervision of these enhanced prototypes in a detector training-free and real-data-free manner, and rectifies post-classification-head representations toward a detector-compatible source-domain visual distribution during inference, thereby improving recognition of novel categories and unseen domains.
As a complementary inference-time strategy, ExRPN recalibrates proposal confidence to improve recall for novel and domain-shifted objects. Experiments on OD-LVIS, OV-LVIS, Objects365, and MSOSB demonstrate that our method achieves SOTA performance and strong category-domain generalization.
% 我们提出了 ExDet，一种面向 ODOVD 的轻量级框架。ExDet 构建于冻结的、基于 CLIP 的两阶段检测器之上，由 Text-Guided Extrapolation（TGE）、Detector-Compatible Rectification（DCR）和 ExRPN 组成。TGE 从文本中推理具备类别感知与领域感知的代理视觉原型；DCR 在这些增强原型的监督下进行独立学习，并在detector training-free and real-data-free 的情况下，在推理阶段将分类头后的表示校正到与检测器兼容的源域视觉分布，从而提升对新类别和未见领域目标的识别能力。作为一种补充性的推理期策略，ExRPN 对 proposal 置信度进行重校准，以提高 novel 和 domain-shifted objects 的召回率。在 OD-LVIS、OV-LVIS、Objects365 和 MSOSB 上的实验表明，本文方法取得了最先进的性能，并展现出良好的类别-领域泛化能力。

{
    \small
    \clearpage
    \bibliographystyle{ACM-Reference-Format}
    \bibliography{reference}
}

\end{document}